\documentclass[11pt]{article}
\usepackage{coling2018}
\usepackage{times}
\usepackage{latexsym}

\usepackage{graphicx}
\usepackage{booktabs} 
\usepackage{amsmath}
\usepackage[justification=centering]{caption}
\usepackage{array}
\usepackage{dblfloatfix}
\usepackage{subcaption}
\usepackage{url}
\usepackage{csvsimple}
\usepackage{algorithm}
\usepackage{algpseudocode}
\usepackage{pgfplots}
\usepackage[utf8]{inputenc}
\usetikzlibrary{pgfplots.groupplots}
\pgfplotsset{compat=1.14}
\newcolumntype{C}[1]{>{\centering\let\newline\\\arraybackslash\hspace{0pt}}m{#1}}

\title{Aff2Vec: Affect--Enriched Distributional Word Representations}

\author{Sopan Khosla, Niyati Chhaya and Kushal Chawla\\
 Big Data Experience Lab\\
 Adobe Research\\
  {\tt \{skhosla,nchhaya,kchawla\}@adobe.com} \\}

\date{}
\begin{document}
\maketitle
\begin{abstract}
Human communication includes information, opinions, and reactions. Reactions are often captured by the affective-messages in written as well as verbal communications. While there has been work in affect modeling and to some extent affective content generation, the area of affective word distributions in not well studied. Synsets and lexica capture semantic relationships across words. These models however lack in encoding affective or emotional word interpretations. Our proposed model, Aff2Vec provides a method for enriched word embeddings that are representative of affective interpretations of words. Aff2Vec outperforms the state--of--the--art in intrinsic word-similarity tasks. Further, the use of Aff2Vec representations outperforms baseline embeddings in downstream natural language understanding tasks including sentiment analysis, personality detection, and frustration prediction. 
\end{abstract}

\section{Introduction}
 \blfootnote{
    %
    \hspace{-0.65cm}  
    This work is licensed under a Creative Commons 
    Attribution 4.0 International License.
    License details:
    \url{http://creativecommons.org/licenses/by/4.0/}
}
Affect refers to the experience of a feeling or emotion~\cite{Scherer2010,Picard1997}. This definition includes emotions, sentiments, personality, and moods. The importance of affect analysis in human communication and interactions has been discussed by Picard \shortcite{Picard1997}. Historically, affective computing has focused on studying human communication and reactions through multi-modal data gathered through various sensors. The study of human affect from text and other published content is an important topic in language understanding. Word correlation with social and psychological processes is discussed by Pennebaker~\shortcite{Pennebaker2011}. Preotiuc-Pietro et al.~\shortcite{perspara17nlpcss} studied personality and psycho-demographic preferences through Facebook and Twitter content. Sentiment analysis in Twitter with a detailed discussion on human affect~\cite{rosenthal2017semeval} and affect analysis in poetry~\cite{Kao2012ACA} have also been explored. Human communication not only contains semantic and syntactic information but also reflects the psychological and emotional states. Examples include the use of opinion and emotion words~\cite{AffectLM}. The analysis of affect in interpersonal communication such as emails, chats, and longer written articles is necessary for various applications including the study of consumer behavior and psychology, understanding audiences and opinions in computational social science, and more recently for dialogue systems and conversational agents. This is a open research space today.

Traditional natural language understanding systems rely on statistical language modeling and semantic word distributions such as WORDNET~\cite{miller1995wordnet} to understand relationships across different words. There has been a resurgence of research efforts towards creating word distributions that capture multi-dimensional word semantics ~\cite{mikolov2013efficient,pennington2014glove}. Sedoc et al.~\shortcite{affnorms17eacl} introduce the notion of affect features in word distributions but their approach is limited to creating enriched representations, and no comments on the utility of the new word distribution is presented. Beyond word-semantics, deep learning research in natural language understanding, is focused towards sentence representations using encoder-decoder models ~\cite{Ahn2016}, integrating symbolic knowledge to language models~\cite{Vinyals2015}, and some recent works in augmenting neural language modeling with affective information to emotive text generation~\cite{AffectLM}. These works however do not introduce distributional affective word representations that not only reflect affective content but are also superior for related downstream natural language tasks such as sentiment analysis and personality detection.\\
\begin{figure*}[!t]
\centering
		\begin{subfigure}[b]{0.30\textwidth}
                \begin{center}\includegraphics[scale=0.35]{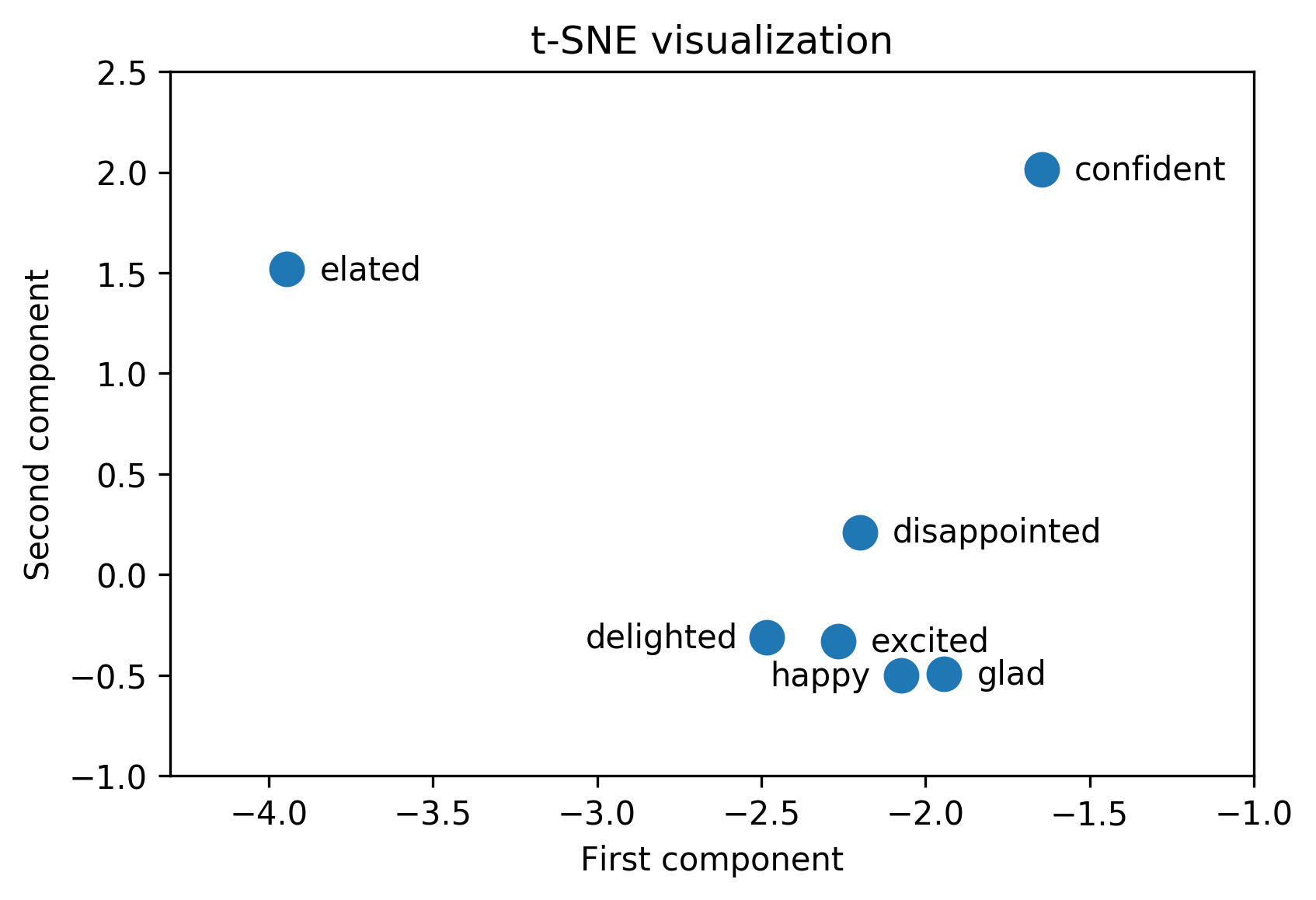}\end{center}
                \caption{GloVe}
        \end{subfigure}
        \begin{subfigure}[b]{0.30\textwidth}
                \begin{center}\includegraphics[scale=0.35]{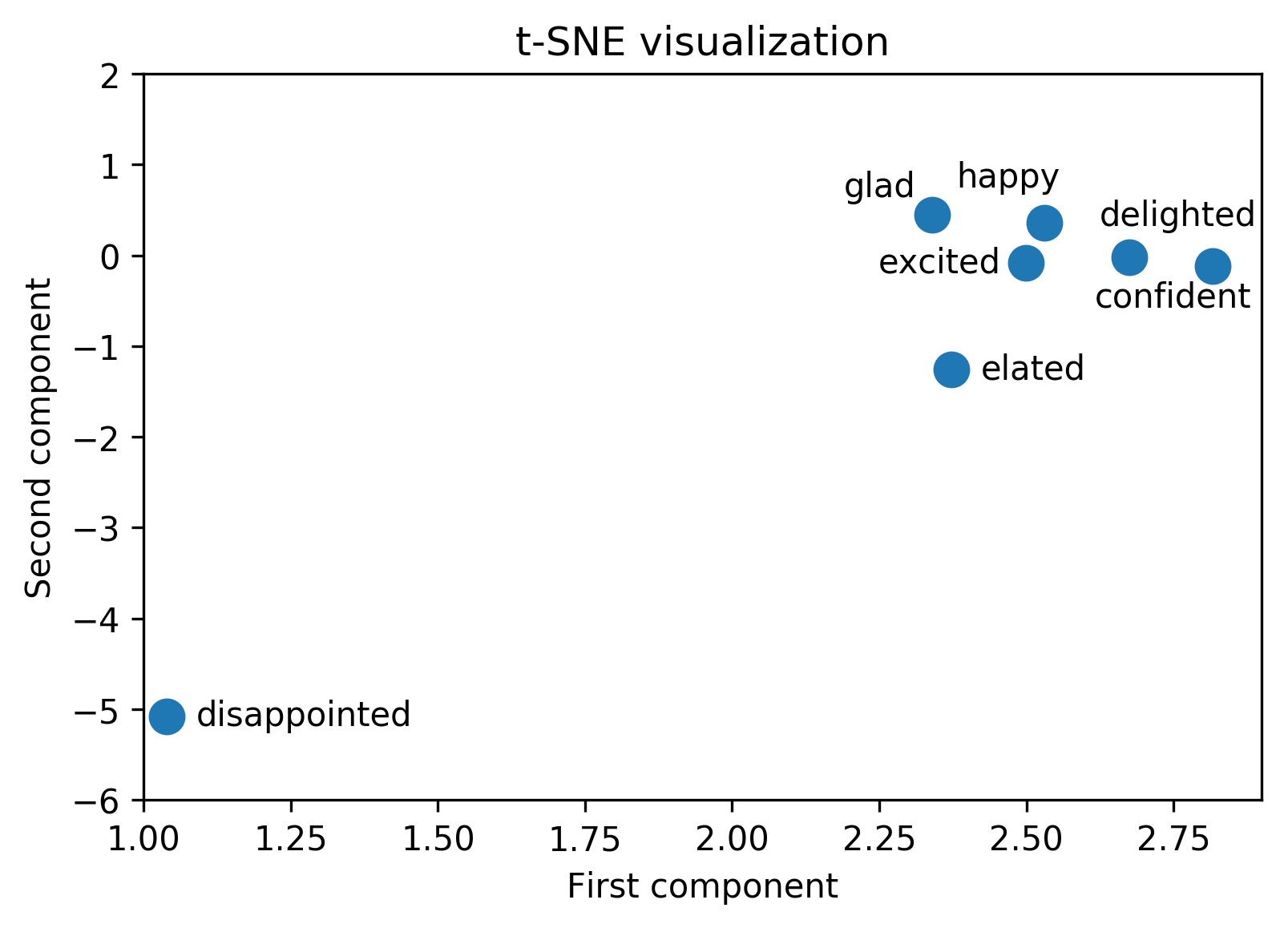}\end{center}
                \caption{GloVe$\oplus$Affect}
        \end{subfigure}%
        \begin{subfigure}[b]{0.30\textwidth}
                \begin{center}\includegraphics[scale=0.35]{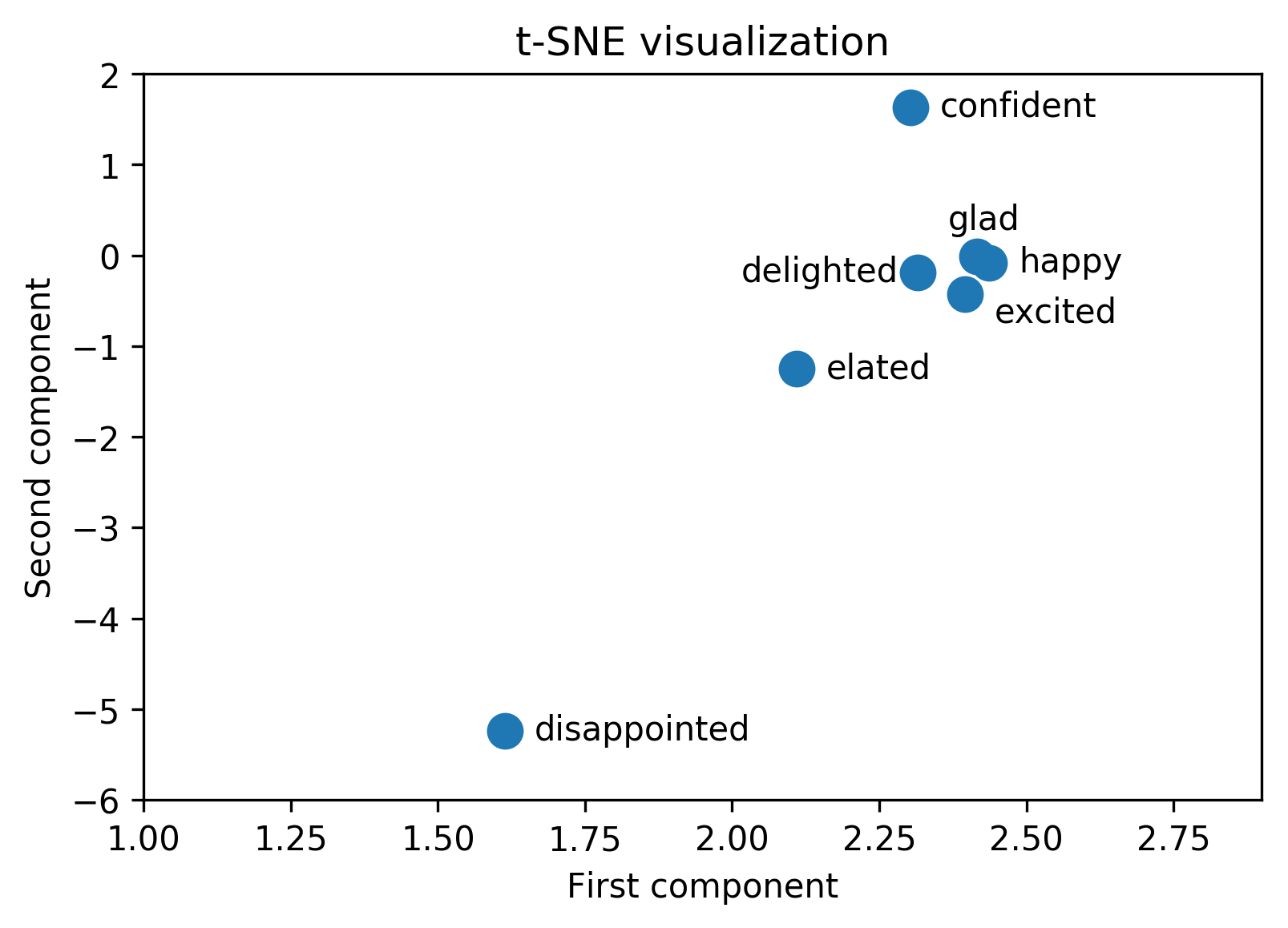}\end{center}
                \caption{GloVe + Counterfitting $\oplus$ Affect}
                     \end{subfigure}%
       \caption{t-SNE for significant affect words: The graphs show the distribution of sample words from Sedoc et al~\shortcite{affnorms17eacl}. The variance in the visualization illustrates the perturbation introduced by distributional schemes discussed in this paper. Vanilla GloVe embeddings show `disappointed' near `delighted', while these are separated in the $\oplus$Affect representations.}
  \label{fig:tsne}  
\end{figure*}
We introduce Aff2Vec, affect-enriched word distributions trained on lexical resources coupled with semantic word distributions. Aff2Vec captures opinions and affect information in the representation using post-processing approaches. Figure~\ref{fig:tsne} illustrates how Aff2Vec captures affective relationships using a t-SNE visualization of the word space. Aff2Vec can be trained using any affect space, we focus on the Valence--Arousal--Dominance dimensions but the approach is generalizable to other space. Our experiments show that Aff2Vec out performs vanilla embedding spaces for both intrinsic word--similarity tasks as well as extrinsic natural language applications. Main contributions of this paper include:\\\\
\textbf{Aff2Vec:} Affect-enriched word representations using post-processing techniques. We show that Aff2Vec outperforms the state-of-the-art in both intrinsic word similarity metrics as well as downstream natural language tasks including Sentiment analysis, Personality detection, and Frustration detection in interpersonal communication.\\
\textbf{ENRON-FFP Dataset:} We introduce the ENRON-FFP Email dataset with Frustration, Formality, and Politeness tags gathered using a crowd-sourced human perception study.\\
The remainder of the paper is organized as follows. The prior art for enriched word distributions is discussed in Section \ref{sec:relatedWork}. Aff2Vec is introduced in section \ref{sec:Affectmod}. We present the crowd-sourcing study for the ENRON-FFP Dataset in section \ref{sec:amt} and section \ref{sec:experiments} discusses the experimental setup. Section \ref{sec:results} presents the evaluation of Aff2Vec for various intrinsic and extrinsic tasks. A discussion on the distributional word representations is presented in section \ref{sec:discussion} before concluding in section \ref{sec:conclusion}.
\section{Related Work}
\label{sec:relatedWork}
The use of lexical semantic information (lexical resources) to improve distributional representations is recent. Methods like \cite{yu2014improving,xu2014rc,bian2014knowledge,kiela2015specializing} achieve this by using word similarity and relational knowledge to modify the prior or add a regularization term. We call such methods `pre-training methods', as they alter the training procedure for word representations. Such methods require a change in the loss function while training the embeddings, hence are computationally expensive.

The other set of word distribution enhancements are done post-training. These methods aim to include external information using normalizations and modifications to the vanilla word distributions. Methods such as Retrofitting ~\cite{faruqui2014retrofitting} which tries to drag similar words closer together, where notion of similarity is taken from word relation knowledge found in semantic lexica (e.g. WordNet) fall in this category. Counterfitting ~\cite{mrkvsic2016counter} on the other hand initiates from SimLex-999 tuned embeddings, injects antonyms and synonym constraints to improve word representations. This paper introduces post-training techniques on vanilla, retrofitted, and counterfitted embeddings to include affective information in the distributions. Our work falls in the post-training category, hence no direct comparison with the pre-trained approaches is presented in this paper.

Recent work has explored approaches to adapt general-purpose lexica for specific contexts and affects. Studies have recognized the limited applicability of general purpose lexica such as ANEW~\cite{bradley1999affective} to identify affect in verbs and adverbs, as they focus heavily on adjectives. Recognizing that general-purpose lexica often detect sentiment which is incongruous with context, Ribeiro et al.~\shortcite{ribeiro2016sentibench} proposed a sentiment damping method which  utilizes the average sentiment strength over a document to damp any abnormality in the derived sentiment strength. Similarly, Blitzer et al.~\shortcite{blitzer2007biographies} argued that words like `predictable' induced a negative connotation in book reviews, while `must-read' implied a highly positive sentiment. This paper doesn't focus on building yet another affect lexicon but studies the consequences of including affect information in distributional word representations that aim at defining relational relationships across all words in large contexts and vocabularies.

Automatic expansion of affect ratings has been approached with the intuition that words closer in the distributional space would have similar ratings \cite{recchia2015reproducing,palogiannidi2015valence,vankrunkelsven2015predicting,koper2016automatically}. Recent work by Sedoc et al.\shortcite{affnorms17eacl} uses Signed Spectral Clustering to differentiate between words which are contextually similar but display opposite affect. Whereas \cite{wang2016community} uses a graph--based method inspired by label propagation. While our approach follows the nature of the task defined in Sedoc et al., we propose a generalized method to enrich content with affective information. They focus on distinguishing the polarities. Our method incorporates both semantic and affect information hence creating embeddings that can also be used for semantic similarity tasks. Note that Sedoc et al. do not include any semantic information in their modeling.
\section{Aff2Vec: Affect--enriched Word Distributions} 
\label{sec:Affectmod}
Aff2Vec aims at incorporating affective information in word representations. We leverage the Warriner's lexicon \cite{warriner2013norms} in the Valence--Arousal--Dominance space for this work. The proposed work is generalizable to other affect spaces.\footnote{Experiments on other spaces are reported in the supplement.} This section presents two approaches for affect--enrichment of word distributions.

\textbf{Warriner's lexicon:} This is a affect lexicon with $13915$ english words. It contains real-valued scored for Valence, Arousal, and Dominance (VAD) on a scale of $1-9$ each. $1$, $5$, $9$ correspond to the low, moderate (i.e. neutral), and high values for each dimension respectively.  The lexicon does not contain common English words such as stop words and proper nouns. For such out--of--dictionary words we assume a neutral affect vector~$\vec{a}=[5,5,5]$.
\subsection{Affect-APPEND}
Consider word embeddings $W$, the aim is to introduce affective information to this space using the affect embedding space, $A$. The word vectors $W$, with dimension $D$ are concatenated with affect vectors $A$ with dimension $F$, thus resulting in a $D+F$ dimensional enriched representation. The process for this concatenation is described here:\\ 
1. Normalize word vector $W$ and affect vector $A$ using their L2-Norms (Equation~\ref{eq:reg}, \ref{eq:regav}). This reduces the individual vectors to unit-length.  
\begin{equation}
x_i = \dfrac{x_i}{\sqrt{\sum_{k = 1}^{D} x_{ik}^2}} ~~\forall x_i \in W, ~~~~a_i = \dfrac{a_i}{\sqrt{\sum_{k = 1}^{F} a_{ik}^2}} ~~\forall a_i \in A
\label{eq:reg}
\end{equation}
2. Concatenate the regularized word vectors $x_i$ with regularized affect vectors $a_i$.
\begin{equation}
WA(w) = W(w) \oplus A(w) 
\label{eq:append}
\end{equation}
3. Standardize ($1$ variance, $0$ mean) the $D+F$ dimensional embeddings to achieve uniform distribution.
\begin{equation}
y_i = \dfrac{y_i - \mu}{\sigma} ~~~~ \forall y_i \in WA
\label{eq:normalize}
\end{equation}
where $\mu$ and $\sigma$ represent the mean and standard deviation respectively.\\
4. The enriched space $WA$ is then reduced to original $D$ dimensional vector. We use Principal Component Analysis for the dimensionality reduction.
\subsection{Affect-STRENGTH} 
In this approach, the strength  in the antonym/synonym relationships of the words is incorporated in the word distribution space. Hence, we leverage the Retrofitted Word Embeddings for this approach\cite{faruqui2014retrofitting} \footnote{https://github.com/mfaruqui/retrofitting}.\\\\
\textbf{Retrofitting}: Let $V = \{w_1, w_2, w_3,..., w_n\}$ be a vocabulary and $\Omega$ be an ontology which encodes semantic relations between words present in $V$ (e.g. WORDNET). This ontology $\Omega$ is represented as an undirected graph $(V,E)$ with words as vertices and $(w_i, w_j)$ as edges indicating the semantic relationship of interest. Each word $w_i \in V$ is represented as a vector representation $\hat q_i \in R^d$ learnt using a data--driven approach (e.g. Word2Vec or GloVe) where $d$ is the length of the word vectors.

Let $\hat Q$ be the matrix collection of these vector representations. The objective is to learn the matrix $ Q = (q_1,..., q_n) $ such that the word vectors ($q_i$) are both close to their counterparts in $\hat Q$ and to adjacent vertices in $\Omega$. The distance between a pair of vectors is defined to be Euclidean, hence the objective function for minimization is
\begin{equation}
\resizebox{0.5\textwidth}{!}{
$\Psi(Q) = \sum_{i=1}^{n} {\Bigg[   \alpha_i {\|q_i - \hat q_i\|}^2 + \sum_{(i,j) \in E} {\beta_{ij}{\|q_i - q_j\|}^2}    \Bigg] }
\label{fig:retrofit}$
}
\end{equation}
where $\alpha$ and $\beta$ are hyper parameters and control the relative strengths of the two associations. $\Psi$ is a convex function in $Q$ and its global optimal solution can be found by using an iterative update method. By setting $\frac{\partial \Psi(Q)}{\partial q_i} = 0$, the online updates are as follows:
\begin{equation}
q_i = \frac{\sum_{j:(i,j) \in E} {\beta_{ij}q_j + \alpha_i\hat q_i}}{\sum_{j:(i,j) \in E} {\beta_{ij} + \alpha_i}}
\label{fig:retrofit_update}
\end{equation}

We propose two ways to modify $\beta_{ij}$ in equation \ref{fig:retrofit} in order to incorporate affective strength in the edge weights connecting two retrofitted vectors to each other.\\
\textbf{Affect-cStrength}: In this approach, the affective strength is considered as a function of all $F$ affect dimensions.
\begin{equation}
S(w_i, w_j) = 1 - \dfrac{\|a_{i} - a_{j}\|}{\sqrt{\sum_{f=1}^{F}{max\_dist_f^{2}}}}
\end{equation}
where $a_i$ and $a_j$ are $F$ dimensional vectors in $A$ and $max\_dist_f$ is defined as the maximum possible distance between two vectors in $f^{th}$ dimension ($= 9.0 - 1.0 = 8.0$ for VAD dimensions).\\\\
\textbf{Affect-iStrength}: Here, each dimension is treated individually. For every dimension $f$ in $A$, we add an edge between neighbors in the Ontology $\Omega$ where the strength of that edge is given by $S_{f}(w_i, w_j)$:

\begin{equation}
S_{f}(w_i, w_j) = 
1 - \dfrac{|a_{if} - a_{jf}|}{max\_dist_{f}}, ~~~~ S(w_i, w_j) = \sum_{f=1}^{F}{S_{f}(w_i, w_j)}
\end{equation}
$\beta_{ij}$ from equation \ref{fig:retrofit_update} is normalized with this strength function as 
$\beta_{ij} = \beta_{ij} * S(w_i, w_j)$, 
where $S(w_i,w_j)$ is defined by either Affect-cStrength or Affect-iStrength. \\\\
\section{Dataset: ENRON-FFP}
\label{sec:amt}
\begin{table}[h]
\begin{center}
\scalebox{0.75}{

\begin{minipage}{.45\textwidth}
\caption{Enron-FFP Dataset Description}
\begin{tabular}{lr}
\hline
\textbf{Property} & \textbf{Value} \\ \hline
Total number of emails (Main Experiment) &960 \\
Total number of emails (Pilot Experiment) &90 \\ \hline
Min. sentences per email & 1\\
Max. sentences per email & 17\\
Average email size (no. of sentences) & 4.22 \\ \hline
Average number of words per email & 77.5 \\ \hline
\end{tabular}
\label{tab:data_descFFP}
\end{minipage}
\hspace{2cm}
\begin{minipage}{.60\textwidth}
\caption{Datasets for Intrinsic Evaluation}
\begin{tabular}{lr}
\hline
\textbf{Dataset} & \textbf{\# Word-Pairs} \\ \hline
Word Similarity (\textbf{WS})~\cite{finkelstein2001placing} &353 \\
\textbf{RG}-65~\cite{rubenstein1965contextual} &65 \\
\textbf{MEN}~\cite{bruni2012distributional} & 3000\\
Miller-Charles (\textbf{MC})~\cite{miller1991contextual} & 30\\
\textbf{RW}~\cite{luong2013better} & 2034 \\
\textbf{SCWS}~\cite{huang2012improving} & 2023 \\ 
SimLex-999 (\textbf{SL})~\cite{hill2016simlex} & 999 \\
SimVerb-3500 (\textbf{SV})~\cite{gerz2016simverb} & 3500 \\\hline
\end{tabular}
\label{tab:intrinsic_sets}
\end{minipage}
}
\end{center}
\end{table}

\begin{table*}[h]
\vspace{3mm}
\caption{Example emails with varying inter-annotator agreements.}
\vspace{-2mm}
\scalebox{0.7}{
\begin{tabular}{p{4.5cm}p{12.5cm}p{4.5cm}}
\hline
\textbf{Affect Dimension} & \textbf{Example} & \textbf{Annotations}  \\ \hline
Frustration: Low Agreement &  See highlighted portion.  We should throw this back at Davis next time he
points the finger. & (-1, -1, 0, 0, -2, -2, 0, 0, -2, 0) \\ \hline
Frustration: High Agreement &  Please see announcement below.
Pilar, Linda, India and Deb, please forward to all of your people.
Thanks in advance, adr & (0, 0, 0, 0, 0, 0, 0, 0, 0, 0) \\ \hline
Formality: Low Agreement & I talked with the same reporters yesterday (with Palmer and Shapro).
Any other information that you can supply Gary would be appreciated.  Steve, did Gary A. get your original as the CAISO turns email? GAC & (0, 0, -1, 1, 1, 1, 0, -1, -2, -1) \\ \hline
Politeness: High Agreement & John, This looks fine from a legal perspective.  Everything in it is either already in the public domain or otherwise non-proprietary. Kind regards, Dan & (1, 1, 1, 1, 1, 1, 1, 1, 2, 1) \\ \hline
\end{tabular}
\label{tab:examples_data}
}
\end{table*}
We introduce an email dataset, a subset of the ENRON data~\cite{cohen2009enron}, with tags about interpersonal communication traits, namely, Formality, Politeness, and Frustration. The dataset provides the text, user information, as well as the network information for email exchanges between Enron employees.\\\\
\textbf{Human Perceptions and Definitions}: \textit{Tone} or affects such as frustration and politeness are highly subjective measures. In this work, we do not attempt to introduce or standardize an accurate definition for frustration (or formality and politeness). Instead, we assume that these are defined by human perception, and each individual may differ in their understanding of these metrics. This approach of using untrained human judgments has been used in prior studies of pragmatics in text data~\cite{pavlick2016empirical,danescu2013computational} and is a recommended way of gathering gold-standard annotations~\cite{sigley1997text}. The tagged data is then used to predict the formality, frustration, and politeness tags using Aff2Vec embeddings.\\\\
\textbf{Dataset Annotation}:
We conducted a crowd sourced experiment using Amazon's Mechanical Turk\footnote{https://www.mturk.com/mturk/welcome}. The analysis presented in this section is based on $1050$ emails that were tagged across multiple experiments\footnote{Link to the annotated ENRON-FFP dataset: https://bit.ly/2IAxPab}. Table \ref{tab:data_descFFP} provides an overview of the data statistics of the annotated data. We follow the annotation protocol of the Likert Scale~\cite{allen2007likert} for all three dimensions. Each email is considered as a single data point and only the text in the email body is provided for tagging. Frustration is tagged on a 3 point scale with neutral being equated to `not frustrated'; `frustrated' and `very frustrated' are marked with $-1$ and $-2$ respectively. Formality and politeness follow a $5$ point scale from $-2$ to $+2$ where both extremes mark the higher degree of presence and absence of the respective dimension. Table \ref{tab:examples_data} shows some example emails from the dataset.\\\\
\textbf{Inter-annotator Agreement}: To measure whether the individual intuition of the affect dimensions is consistent with other annotators' judgment, we use interclass correlation\footnote{We report the average raters absolute agreement (ICC1k) using the psych package in R.} to quantify the ordinal ratings. This measure accounts for the fact that we may have different group of annotators for each data point. Each data point has $10$ distinct annotations. Agreements reported for $3$ class and $5$ class annotations $0.506 \pm 0.05$, $0.73 \pm 0.02$, and $0.64 \pm 0.03$ for frustration, formality, and politeness respectively. The agreement measures are similar to those reported for other such psycholinguistic tagging tasks.
\section{Experiments}
\label{sec:experiments}
\vspace{-2mm}
Two sets of experiments are presented to evaluate Aff2Vec embeddings\footnote{Link to the Aff2Vec word embeddings: https://bit.ly/2HGohsO} - Intrinsic evaluation using word similarity tasks and extrinsic evaluation using multiple NLP applications. We focus on $3$ vanilla word embeddings: GloVe~\cite{pennington2014glove}, Word2Vec-SkipGram\footnote{https://code.google.com/archive/p/word2vec/}~\cite{NIPS2013_5021}, and Paragram-SL999~\cite{wieting2015paraphrase}. The vocabulary and embeddings used in our experiments resonate with the experimental setup by Mrk{\v{s}}i{\'c} et al.\shortcite{mrkvsic2016counter} ($76427$ words).
\vspace{-1mm}
\subsection{Intrinsic Evaluation}
Word similarity is a standard task used to evaluate embeddings~\cite{mrkvsic2016counter,faruqui2014retrofitting,bollegala2016joint}. In this paper, we evaluate the embeddings on benchmark datasets given in Table~\ref{tab:intrinsic_sets}.

We report the Spearman's rank correlation coefficient between rankings produced by our model (based on cosine similarity of the pair of words) against the benchmark human rankings for each dataset.
\begin{table*}[!t]
\caption{Intrinsic Evaluation: Word Similarity--We report the Spearman's correlation coefficient~($\rho$).  The results show that Aff2Vec variants improve performance consistently.}
\label{tab:intrinsic}
\vspace{-2mm}
\begin{center}
\resizebox{0.75\textwidth}{!}{%
\begin{tabular}{l c c c c c c c c}
\hline
\textbf{Model} & \multicolumn{8}{c}{\textbf{Word Similarity}}\\ \hline
& \textbf{SL} & \textbf{SV} & \textbf{WS} & \textbf{RG} & \textbf{RW} & \textbf{SCWS} & \textbf{MC} & \textbf{MEN} \\ \hline
\textbf{GloVe} & 0.41& 0.28& 0.74 & 0.77 & 0.54  & 0.64 & 0.80  & 0.80  \\ 
$\oplus$ Affect & 0.49 & 0.39  & \textbf{0.77}  &	0.79 &	0.59  &	0.67  &	0.80 &	0.84 \\ 
+ Retrofitting & 0.53  & 0.37  & 0.73   & 0.81 & 0.52  & 0.66  & 0.82 & 0.82\\ 
+ Retrofitting $\ast$ c-strength &0.53 &0.36 &0.74 &0.81 &0.52 &0.66 &0.82 &0.82\\
+ Retrofitting $\ast$ i-strength &0.56 &0.38 &0.64 &0.80 &0.44 &0.62 &0.80 &0.78\\
+ Retrofitting $\oplus$ Affect & 0.60  & 0.46 & 0.76 & 0.81  & \textbf{0.61} & \textbf{0.69} & 0.81 & \textbf{0.85}\\ 
+ Counterfitting & 0.58  & 0.47  & 0.65 &	0.80 &	0.56  &	0.61  &	0.78 &	0.77  \\ 
+ Counterfitting $\oplus$ Affect & \textbf{0.62} & \textbf{0.53} & 0.70  &	\textbf{0.84}  & \textbf{0.61} & 0.64 & \textbf{0.84} &0.80 \\ \hline
\textbf{Word2Vec} & 0.45 &	0.36&	0.70 &	0.76&0.59 &	0.67 &	0.80&	0.78\\
$\oplus$ Affect & 0.49&	0.42 &	0.67 &	0.81 &	0.59&	0.66 &	0.85&	0.79 \\
+ Retrofitting & 0.55& 0.45  &	\textbf{0.74}	&0.82	&\textbf{0.62}	&\textbf{0.70}	&0.83	&0.80  \\
+ Retrofitting $\ast$ c-strength &0.55 &0.44 &0.73 &0.82 &0.62 &\textbf{0.70} &0.83 &0.80 \\
+ Retrofitting $\ast$ i-strength &0.58 &0.47 &0.71 &0.83 &0.57 &0.69 &0.85 &0.80 \\
+ Retrofitting $\oplus$ Affect & 0.59&	0.49 &	0.71 &	\textbf{0.84}&	\textbf{0.62}&	\textbf{0.70} &	\textbf{0.86 }&	\textbf{0.82} \\
+ Counterfitting & 0.56 &	0.51 &	0.66 &	0.75 &	0.61&	0.64 &0.75 &	0.73 \\
+ Counterfitting $\oplus$ Affect & \textbf{0.60} &\textbf{0.54}&	0.64 &	0.82&	0.60 &	0.64&	0.82&	0.76 \\ \hline
\textbf{Paragram} &0.69 &	0.54 &\textbf{0.73}&0.78 &0.59 &0.68&	0.80&	0.78  \\
$\oplus$ Affect & 0.71  & 0.59  & 0.70 & 0.77 & \textbf{0.60}  & 0.67 	& 0.76  & \textbf{0.79}  \\ 
+ Retrofitting & 0.68  & 0.55  & \textbf{0.73} & 0.79 & 0.59  & 0.68 & 0.81  & 0.78 \\
+ Retrofitting $\ast$ c-strength &0.69 &0.55 &\textbf{0.73} &0.79 &0.59 &\textbf{0.69} &0.81 & 0.78\\
+ Retrofitting $\ast$ i-strength &0.68 &0.56 &0.71 &0.80 &0.58 &0.68 &\textbf{0.84} & 0.77\\
+ Retrofitting $\oplus$ Affect & 0.71  & 0.58 & 0.70 & 0.80  & 0.59  & 0.67 & 0.78  & \textbf{0.79} \\ 
+ Counterfitting & 0.74 & 0.63  & 0.69& \textbf{0.81} & \textbf{0.60} & 0.66  & 0.82 & 0.74 \\
+ Counterfitting $\oplus$ Affect & \textbf{0.75} & \textbf{0.66} & 0.68 & \textbf{0.81} & \textbf{0.60} & 0.65 & 0.82 & 0.76  \\ \hline
\end{tabular}
}
\end{center}
\vspace{-2mm}
\end{table*}
\begin{table*}[!ht]
\centering
\caption{Extrinsic Evaluation: Results for FFP--Prediction, Personality Detection, Sentiment Analysis and WASSA Emotional Intensity task for Aff2Vec variants for GloVe and Word2Vec embeddings. We report the Mean Squared Error (MSE) for FFP--Prediction, Accuracy (\% ACC) for Personality Detection and Sentiment Analysis (SA) and Person's $\rho$ for the WASSA Emo-Int Task (EMO-INT)}
\resizebox{\textwidth}{!}{%
\begin{tabular}{lccc@{\hskip 0.3in}ccccc@{\hskip 0.3in}ccccc}
\hline
\textbf{Model} & \multicolumn{3}{c@{\hskip 0.3in}}{\textbf{FFP-Prediction}} & \multicolumn{5}{c@{\hskip 0.3in}}{\textbf{Personality Detection}} & \textbf{SA} & \multicolumn{4}{c}{\textbf{EMO-INT}}\\ \hline
 &  \multicolumn{3}{c@{\hskip 0.3in}}{\textbf{MSE }($X 10^{-3}$)} & \multicolumn{5}{c@{\hskip 0.3in}}{\textbf{Acc. ($\%$)}} & \textbf{Acc.} ($\%$) & \multicolumn{4}{c}{\textbf{Pearson's $\rho$} ($X 10^{-2}$)} \\ \hline
& \textbf{FOR} & \textbf{FRU} & \textbf{POL} & \textbf{EXT}& \textbf{NEU}& \textbf{AGR}& \textbf{CON}& \textbf{OPEN} & \textbf{DAN} & \textbf{ANG} &\textbf{FEA} &\textbf{JOY} &\textbf{SAD}\\ \hline
\textbf{GloVe} & 27.59 & 32.40 & 21.89 & \textbf{56.08} & 55.25 & 56.06 & \textbf{57.32} & 59.14 & 83.1 &70.98 &71.19 &65.85 &73.30\\
$\oplus$ Affect & 27.72 & 28.76 & 22.02&  51.47 & 57.41 & 56.09 & 55.06 & \textbf{62.08} & 84.3&70.91 &71.72 &66.26 &\textbf{73.58} \\
+ Retrofitting &27.44 & 29.35 & 21.75& 55.79 & 59.67 & 55.59 & 56.89 & 59.67 & 82.7&72.10 &71.86 &\textbf{67.11} & 73.14\\ 
+ Retrofitting $\oplus$ Affect& 28.33 & \textbf{27.91} & 22.24 &55.01 & 56.43 & \textbf{57.48} & 53.04 & 61.12 & 83.7&\textbf{72.38} &\textbf{72.53} &66.29 &72.76 \\
+ Counterfitting &\textbf{25.66} & 29.20 & 22.90& 55.11 & 58.32 & 55.41 & 53.89 & 60.36 & 84.2 &70.45 &68.95 &65.27 &72.63\\
+ Counterfitting $\oplus$ Affect &28.89 & 32.46 & \textbf{21.64}& 52.12 & \textbf{60.03} & 56.53 & 54.93 & 59.51 & \textbf{84.4}&70.20 &70.43 &65.81 &72.37 \\ \hline
\textbf{Word2Vec} &25.86 & 27.88 & 21.56& \textbf{56.08}  & 58.19 & 56.59 & 55.18 & 61.41 & 83.3 &68.86 &71.24 &65.23 &72.60 \\
$\oplus$ Affect & 25.39 & 28.16 & 22.99 & 53.54 & 57.97 & 55.17 & 54.12 & 59.31 & 83.4 & 69.29 &\textbf{71.92} &64.49 &\textbf{72.63} \\
+ Retrofitting&27.81 & 29.05 & 21.85  & 54.33 & 56.65 & \textbf{57.39} & 54.65 & 60.03 & 82.5 &70.12 &71.42 &\textbf{67.96} &72.02 \\
+ Retrofitting $\oplus$ Affect & \textbf{25.08} & \textbf{27.08} & 21.64  &53.74 & \textbf{59.61} & 56.34 & \textbf{56.93} & 59.7 &83.3 & \textbf{70.65} &71.90 &66.36 &72.20 \\
+ Counterfitting & 28.28 & 27.12 & 22.95& 54.55 & 57.61 & 57.09 & 54.1 & 58.5 & 83.3 & 68.64 & 70.13 & 63.36 & 70.67 \\
+ Counterfitting $\oplus$ Affect &27.73 & 29.67 & \textbf{21.52} & 51.28 & 58.86 & 56.66 & 53.22 & \textbf{61.62} & \textbf{83.5} & 69.38 &70.31 &64.94 &71.37 \\ \hline
\textbf{Baselines} &&&&&&&&&&&&&\\
\cite{majumder2017deep} & -- & -- & -- & \textbf{58.09} & 59.38 & 56.71 & 57.30 & \textbf{62.68} & -- &-- &-- &-- &--\\ 
ENRON Trainable & 31.61 & 43.90 & 26.27 & -- & -- & -- & -- & -- & -- &-- &-- &-- &--\\ 
Re(Glove)\cite{yu2017refining} & -- & -- & -- & -- & -- & -- & -- & -- & 82.2 &-- &-- &-- &-- \\ 
Re(w2v)\cite{yu2017refining} & -- & -- & -- & -- & -- & -- & -- & -- & 82.4 &-- &-- &-- &--  \\ \hline
\end{tabular}%
}
\vspace{-2mm}
\label{tab:extrinsic_eval}
\end{table*}
\vspace{-1mm}
\subsection{Extrinsic Evaluation}
\vspace{-1mm}
Although intrinsic tasks are popular, performance of word embeddings on these benchmarks does not reflect directly into the downstream tasks~\cite{chiu2016intrinsic}. \cite{gladkova2016intrinsic,batchkarov2016critique} suggest that intrinsic tasks should not be considered as gold standards but as a tool to improve the model. We test the utility of the Aff2Vec on $4$ distinct natural language understanding tasks:\\
\textbf{Affect Prediction}~(FFP-Prediction): The experiment is to predict the formality, politeness, and frustration in email. We introduce the ENRON-FFP dataset for this task in section~\ref{sec:amt}. A basic CNN model is used for the prediction\footnote{Hyper-parameters and model details are discussed in the supplementary material}. The purpose of this experiment is to evaluate the quality of the embeddings and not necessarily the model architecture. The CNN is hence not optimized for this task. Embeddings trained on the ENRON dataset (ENRON-Trainable) are used as a baseline.\\\\
\textbf{Personality Detection}: This task is to predict human personality from text. The big five personality dimensions~\cite{digman1990personality} are used for this experiment. The $5$ personality dimensions include Extroversion (EXT),
Neurotic-ism (NEU), Agreeableness (AGR), Conscientiousness (CON), and Openness (OPEN). Stream-of-consciousness essay dataset by Pennebaker et al.~\shortcite{pennebaker1999linguistic} contains $2468$ anonymous essays tagged with personality traits of the author. We use this dataset. Majumder et al \shortcite{majumder2017deep} propose a CNN model for this prediction. We use their best results as baseline and report the performance of Aff2Vec on their default implementation\footnote{https://github.com/SenticNet/personality-detection}.\\\\
\textbf{Sentiment Analysis}: The Stanford Sentiment Treebank~(SST)~\cite{socher2013recursive} contains sentiment labels on sentences from movie reviews. This dataset in its binary form is split into training, validation, and test sets with $6920$, $872$, and $1821$ samples, respectively. We report the performance on a Deep Averaging Network (DAN)\footnote{https://github.com/miyyer/dan} \cite{iyyer2015deep} with default parameters on the SST dataset and compare against refined embeddings specifically created for sentiment analysis. Implementation by Yu et al~\shortcite{yu2017refining} is used for the refined embeddings.\footnote{Implementation provided by the authors is used for this experiment.}\\\\
\textbf{Emotion Intensity Task (WASSA)}: WASSA shared task on emotion intensity \cite{W17-5205} requires to determine the intensity of a particular emotion (anger, fear, joy, or sadness) in a tweet. This intensity score can be seen as an approximation of the emotion intensity of the author or as felt by the reader. We train a BiLSTM-CNN--based model for this regression task with embedding dimensions as features.\footnote{Model details are provided as supplementary material.}. Vanilla embeddings are used as a baseline for this experiment.
\subsection{Qualitative Evaluation: Noise@k}
Affect-enriched embeddings perform better as they move semantically similar but affectively dissimilar words away from each other in the vector space. We demonstrate this effect through two measures that capture noise in the neighborhood of a word. \\\\
\textit{Polarity-Noise@k} (PN@k)~\cite{yu2017refining} calculates the number of top $k$ nearest neighbors of a word with opposite polarity for the affect dimension under consideration.\\
\textit{Granular-Noise@k} (GN@k) captures the average difference between a word and its top $k$ nearest neighbors for a particular affect dimension ($f$). 
\begin{equation}
GN_i@k = 
\dfrac{\sum_{j \in kNN_i}{|a_if - a_jf|}}{k}
\end{equation}
where $a_i$, $a_j$ are $F$--dimensional vectors in $A$ and $kNN_i$ denotes the top $k$ nearest neighbors of word $i$. This is done for each word in the affect lexicon.
  
\begin{table*}[!t]
\caption{Polarity-Noise@k~(PN@10) and Granularity-Noise@k~(GN@10) where $k=10$ for GloVe and Word2Vec variants. Note that lower the number, better this qualitative metric.}
\label{tab:noise}
\centering
\resizebox{0.6\textwidth}{!}{%
\begin{tabular}{lccc@{\hskip 0.3in}ccc}
\hline
\textbf{Model} &  \multicolumn{3}{c@{\hskip 0.3in}}{\textbf{PN@10 ($\%$)}} & \multicolumn{3}{c}{\textbf{GN@10 ($X 10^{-2}$)}}\\ \hline
 & \textbf{V} & \textbf{A} & \textbf{D} & \textbf{V} & \textbf{A} & \textbf{D} \\ \hline
\textbf{GloVe} & 23.21 & 22.15 & 27.07 & 83.91 & 79.19 & 74.19 \\ 
$\oplus$ Affect & \textbf{16.46} & 19.65 & \textbf{19.42} & \textbf{72.56} & \textbf{69.00} & 64.02 \\ 
+ Retrofitting & 22.55 & 21.82 & 26.5 & 82.15 & 78.68 & 72.53 \\ 
+ Retrofitting $\ast$ c-strength &22.07 &21.63 &26.14 &80.85 &78.12 &71.86 \\
+ Retrofitting $\ast$ i-strength  &23.05 &21.77 &26.66 &83.14 &78.76 &72.65 \\
+ Retrofitting $\oplus$ Affect & 19.68 & \textbf{18.16} & 22.88 & 73.45 & 71.56 & 66.55 \\
+ Counterfitting & 22.68 & 22.2 & 26.46 & 83.31 & 78.78 & 72.54 \\
+ Counterfitting $\oplus$ Affect & 16.75 & 19.99 & 19.99 & 73.89 & 69.55 & \textbf{63.93} \\ \hline
\textbf{Word2Vec} & 24.66 & 22.19 & 27.41 & 85.81 & 79.23 & 74.25 \\ 
$\oplus$ Affect & 20.62 & \textbf{17.83} & 23.19 & \textbf{74.78} & \textbf{71.64} & 67.32 \\ 
+ Retrofitting & 23.75 & 22.25 & 26.94 & 84.65 & 79.36 & 73.00 \\ 
+ Retrofitting $\ast$ c-strength &23.33 &22.01 &26.58 &83.39 &78.71&72.24  \\
+ Retrofitting $\ast$ i-strength &23.90 &22.30 &27.13 &85.34 &79.46 &73.12 \\
+ Retrofitting $\oplus$ Affect & 20.61 & 18.54 & 23.6 & 75.71 & 72.47 & 67.61 \\ 
+ Counterfitting & 23.47 & 22.48 & 26.72 & 84.62 & 79.14 & 72.29 \\ 
+ Counterfitting $\oplus$ Affect & \textbf{20.34} & 18.17 & \textbf{23.01} & 74.83 & 71.94 & \textbf{66.62} \\ \hline
\textbf{Paragram} & 25.16 & 22.55 & 28.05 & 88.34 & 80.73 & 75.49 \\ 
$\oplus$ Affect & 20.81 & 21.29 & 23.45 &81.83 & 75.27 & 69.79 \\ 
+ Retrofitting & 25.69 & 22.8 & 28.48 & 89.67 & 81.25 & 76.05 \\ 
+ Retrofitting $\ast$ c-strength &25.46 &22.64 &28.22 &89.06 &80.95 &75.58 \\ 
+ Retrofitting $\ast$ i-strength &25.69 &22.84 &28.43 &89.85 &81.26 &75.93 \\
+ Retrofitting $\oplus$ Affect & 23.38 & \textbf{20.34} & 25.99 & 83.17 & 76.51 & 71.83 \\
+ Counterfitting & 24.86 & 22.76 & 27.88 & 88.27 & 80.68 & 75.18 \\ 
+ Counterfitting $\oplus$ Affect & \textbf{20.31} & 21.5 & \textbf{23.03} & \textbf{81.40} & \textbf{75.05} & \textbf{69.10} \\ \hline
\end{tabular}
}
\vspace{-2mm}
\end{table*}
\begin{table*}
\caption{Top-5 NN for `Good' and `Bad' for variants of GloVe, SentiWordNet and Aff2Vec}
\label{tab:nearestneighbours}
\begin{center}
\resizebox{0.95\textwidth}{!}{%
\begin{tabular}{lll}
\hline
\textbf{Model} & \textbf{Good} & \textbf{Bad}\\ \hline
\textbf{GloVe} &[\textit{great}, \textit{nice}, \textit{excellent}, \textit{decent}, \textit{bad}] & [\textit{terrible}, \textit{awful}, \textit{horrible}, \textit{wrong}, \textit{thing}]\\
$\oplus$ Affect &[\textit{great}, \textit{nice}, \textit{excellent}, \textit{decent}, \textit{pretty}] & [\textit{awful}, \textit{terrible}, \textit{horrible}, \textit{wrong}, \textit{crappy}] \\
+ Retrofitting &[\textit{great}, \textit{decent}, \textit{nice}, \textit{excellent}, \textit{pretty}]	&[\textit{wrong}, \textit{awful}, \textit{terrible}, \textit{horrible}, \textit{nasty}] \\
+ Retrofitting $\oplus$ Affect &[\textit{nice}, \textit{great}, \textit{decent}, \textit{excellent}, \textit{pretty}]&	[\textit{awful}, \textit{wrong}, \textit{nasty}, \textit{terrible}, \textit{horrible}] \\ 
+ Counterfitting &[\textit{decent}, \textit{nice}, \textit{optimum}, \textit{presentable}, \textit{exemplary}] & [\textit{rotten}, \textit{shitty}, \textit{horrid}, \textit{naughty}, \textit{lousy}]\\
+ Counterfitting $\oplus$ Affect &[\textit{nice}, \textit{decent}, \textit{optimum}, \textit{presentable}, \textit{dignified}] & [\textit{rotten}, \textit{shitty}, \textit{horrid}, \textit{lousy}, \textit{naughty}] \\ \hline
Senti-WordNet\protect\footnotemark & [\textit{commodity}, \textit{full}, \textit{estimable}, \textit{beneficial}, \textit{adept}]  & [\textit{regretful}, \textit{badly}]\\ \hline
Warriner's Lexicon &[\textit{grandmother}, \textit{healing}, \textit{cheesecake}, \textit{play}, \textit{blissful}] & [\textit{jittery}, \textit{fuss}, \textit{incessant}, \textit{tramp}, \textit{belligerent}] \\ \hline
\end{tabular}
}
\end{center}
\vspace{-2mm}
\end{table*}
\section{Results}
\vspace{-2mm}
\label{sec:results}
All experiments are compared against the vanilla word embeddings, embeddings with counterfitting, and embeddings with retrofitting.

Table \ref{tab:intrinsic} summarizes the results of the \textbf{Intrinsic word--similarity tasks}. For the pre--trained word embeddings, Paragram-SL999 outperformed GloVe and Word2Vec on most metrics. Both retrofitting and counterfitting procedures show better or at par performance on all datasets except for WordSim-353. Addition of affect information to different versions of GloVe consistently improves performance whereas the only significant improvement for Paragram-SL999 variants is observed on the SimLex-999 and SimVerb-3500 datasets. To the best of our knowledge, $\rho=0.74$ reported by \cite{mrkvsic2016counter} represents the current state--of--the--art for SimLex-999 and inclusion of affect information to these embeddings yields higher performance~($\rho = 0.75$). Similarly, for the SimVerb-3500 dataset, Paragram+Counterfitting$\oplus$Affect embeddings beat the state--of--the--art scores\footnote{mentioned at \url{http://people.ds.cam.ac.uk/dsg40/simverb.html}}. Amongst Affect-APPEND and Affect-STRENGTH, Affect-APPEND out performs the rest in most cases for GloVe and Word2vec. However, Affect-STRENGTH variations perform slightly better for the retrofitted Paragram embeddings.

The results for the \textbf{Extrinsic tasks} are reported in Table \ref{tab:extrinsic_eval}. We report the performance for GloVe and Word2Vec with Affect-APPEND variants.\footnote{Results for Paragram are reported in the supplement.} For FFP-Prediction, Affect-APPEND reports the lowest Mean Squared Error for Frustration and Politeness. However, in the case of Formality, the counterfitting variant reports the lowest error. For the personality detection, Affect-APPEND variants report best performance for NEU, AGR, and OPEN classes. For CON, Glove beats the best results in \cite{majumder2017deep}. Evaluation against the Sentiment Analysis(SA) task shows that Affect-APPEND variants report highest accuracies. The final experiment reported here is the WASSA-EmoInt task. Affect-APPEND and retrofit variants out perform the vanilla embeddings.

\begin{figure*}[!t]
\begin{center}
\resizebox{\textwidth}{!}{%
  \begin{tikzpicture}
    \pgfplotsset{footnotesize,samples=10,label style={font=\tiny},
                    tick label style={font=\tiny}}
    \begin{groupplot}[group style = {group size = 3 by 1, horizontal sep = 25pt}, width = 6.2cm, height = 5.0cm, xticklabels from table={v.dat}{X},xtick=data]
        \nextgroupplot[ title = {GN@k for Valence},
            legend style = { column sep = 1pt, legend columns = -1, legend to name = grouplegend,}]
\addplot[blue,thick,mark=square*] table [y=0,x=X]{v.dat};
\addlegendentry{Glove}
\addplot[green,thick,mark=square*] table [y=1,x=X]{v.dat};
\addlegendentry{$\oplus$ Affect}]
\addplot[violet,thick,mark=square*] table [y=2,x=X]{v.dat};
\addlegendentry{+ Retrofitting}]
\addplot[red,thick,mark=square*] table [y=3,x=X]{v.dat};
\addlegendentry{+ Retrofitting $\oplus$ Affect}]
\addplot[orange,thick,mark=square*] table [y=4,x=X]{v.dat};
\addlegendentry{+ Counterfitting}]
\addplot[black,thick,mark=square*] table [y=5,x=X]{v.dat};
\addlegendentry{+ Counterfitting $\oplus$ Affect}]
        \nextgroupplot[title = {GN@k for Arousal},]
\addplot[blue,thick,mark=square*] table [y=0,x=X]{a.dat};
\addplot[green,thick,mark=square*] table [y=1,x=X]{a.dat};
\addplot[violet,thick,mark=square*] table [y=2,x=X]{a.dat};
\addplot[red,thick,mark=square*] table [y=3,x=X]{a.dat};
\addplot[orange,thick,mark=square*] table [y=4,x=X]{a.dat};
\addplot[black,thick,mark=square*] table [y=5,x=X]{a.dat};
        \nextgroupplot[title = {GN@k for Dominance},] 
\addplot[blue,thick,mark=square*] table [y=0,x=X]{d.dat};
\addplot[green,thick,mark=square*] table [y=1,x=X]{d.dat};
\addplot[violet,thick,mark=square*] table [y=2,x=X]{d.dat};
\addplot[red,thick,mark=square*] table [y=3,x=X]{d.dat};
\addplot[orange,thick,mark=square*] table [y=4,x=X]{d.dat};
\addplot[black,thick,mark=square*] table [y=5,x=X]{d.dat};
    \end{groupplot}
    \node at ($(group c2r1) + (0,-3.0cm)$) {\ref{grouplegend}}; 
\end{tikzpicture}
}
\end{center}
\caption{Variation of Granular Noise with different k values for GloVe  and Affect-APPEND variants }
\label{fig:noisek}
\end{figure*}
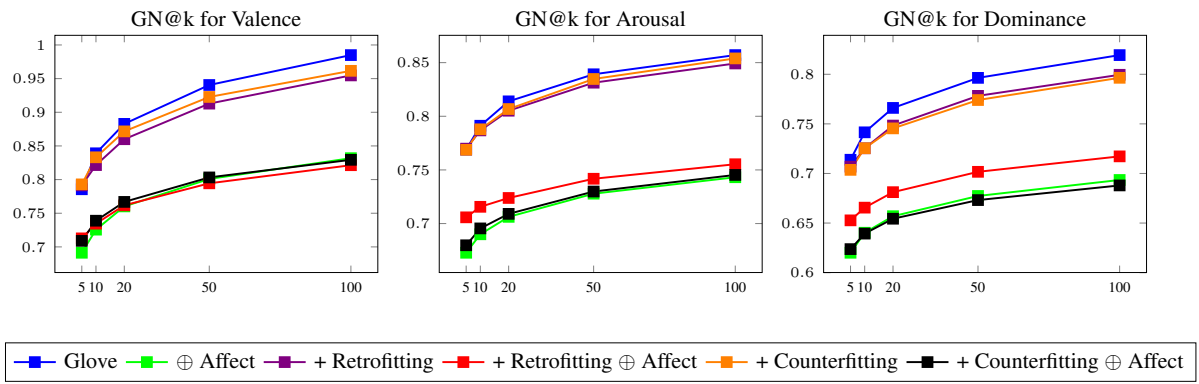

To summarize, the extrinsic evaluation supports the hypothesis that affect--enriched embeddings improve performance for all NLP tasks. Further, the word similarity metrics show that Aff2Vec is not specific to sentiment or affect--related tasks but is at par with accepted embedding quality metrics.\\\\
\textbf{Qualitative Evaluation}: Table \ref{tab:noise} reports the average \textit{Polarity-Noise@10} and \textit{Granular-Noise@10} for GloVe and Word2Vec variants. Note that lower the noise better the performance. The Affect-APPEND report the lowest noise for both cases. This shows that the introduction of affect dimensions in the word distributions intuitively captures psycholinguistic and in particular polarity properties in the vocabulary space. The rate of change of noise with varying $k$ provides insights into (1) how similar are the embedding spaces and (2) how robust are the new representations to the noise - how well is the affect captured in the new embeddings. Figure~\ref{fig:noisek} shows the granular noise@k for Valence, Arousal, and Dominance respectively. Noise@k for the Aff2Vec i.e. the Affect-APPEND variants, specifically, $\oplus$Affect and Couterfitting$\oplus$Affect has lower noise even for a higher $k$. The growth rate for all variants is similar and reduces with an increase in the value of $k$. A similar behavior is observed for Polarity-Noise@k.

\section{Discussion}
\label{sec:discussion}
\vspace{-2mm}
Experiments give an empirical evaluation of the proposed embeddings, none of these provide an insight about the change in the distributional representations of the associated words. Semantic relationship capture the synonym like information. We study how the \textbf{neighborhood} of a certain word changes based on the different word distribution techniques used to create the corresponding representations. Table \ref{tab:nearestneighbours} shows the top five nearest neighbors based on the representations used. While SENTI-Wordnet represents synonyms more than affectively similar words, the affect--enriched embeddings provide a combination of both affective similarity and semantic similarity. The variance in the ranking of words also captures how different schemes capture the intuition of word distributions. Such an analysis can be used to build automated natural language generation and text modification systems with varying objectives.

\section{Conclusion}
\label{sec:conclusion}
We present a novel, simple yet effective method to create affect--enriched word embeddings using affect and semantic lexica. The proposed embeddings outperform the state--of--the--art in benchmark intrinsic evaluations as well as extrinsic applications including sentiment, personality, and affect prediction. We introduce a new human--annotated dataset with formality, politeness, and frustration tags on the publicly available ENRON email data. We are currently exploring the effect of dimension size on the performance of the enriched embeddings as well as the use of Aff2Vec for complex tasks such as text generation. 


\nocite{RotheES16,VulicMRSYK17,SedocGUF16,BojanowskiGJM16}
\bibliographystyle{acl}
\bibliography{coling2018}

\appendix

\end{document}